# The Outputs of Large Language Models are Meaningless[*]

Anandi Hattiangadi & Anders J. Schoubye

*September 22, 2025*

**Abstract**: In this paper, we offer a simple argument for the conclusion that the outputs of large language models (LLMs), such as OpenAI's ChatGPT and Google's Gemini, are meaningless. Our argument is based on two key premises: (a) that certain kinds of intentions are needed in order for LLMs' outputs to have literal meanings, and (b) that LLMs cannot plausibly have the right kinds of intentions. We defend this argument from various types of responses, for example, the semantic externalist argument that deference can be assumed to take the place of intentions and the semantic internalist argument that meanings can be defined purely in terms of intrinsic relations between concepts, such as conceptual roles. We conclude the paper by discussing why, even if our argument is sound, the outputs of LLMs nevertheless seem meaningful and can be used to acquire true beliefs and even knowledge.

---

[*] Earlier versions of this paper have been presented at the CLLAM seminar, Stockholm University, at the Higher Seminar in Theoretical Philosophy, at Uppsala University, and at the Conference: Externalism, Metasemantics and the Interpretatation of AI, arranged by the Institute of Philosophy, University of London. We are grateful to the participants of these events for helpful comments and discussion. Thanks also to Gary Ostertag and Andreas Stokke for further discussion. Hattiangadi's research towards this paper has been generously funded by the MAW Wallenberg foundation for the project 'Cogito Machina: Investigating the Emergence of Artificial General Intelligence' (2025-2028).

# 1   Introduction

It is unquestionable that the large language models (LLMs) that currently power chatbots such as Open AI's ChatGPT, Google's Gemini, and many others, are highly proficient in generating grammatical, fluent and informative pieces of text. However, a fairly simple argument can be constructed with the conclusion that the outputs of LLMs are fundamentally meaningless. Given the unquestionable utility of LLMs, this might seem rather implausible. Nevertheless, the aim of this paper is to argue that the conclusion of this argument is in fact correct, but ultimately also much less implausible than it might at first seem.

# 2   The Argument

First, there are many reasons to believe that speakers' intentions are necessary to determine the meanings of the expressions of a natural language on concrete occasions of use. In part, this is because there is a multitude of cases in which a single expression form can be used to express distinct concepts, refer to different entities, and express different truth-evaluable contents—i.e. *propositions*—on different occasions.[1] In such cases, it is evident that the speaker's communicative intention is needed to determine which of several candidate interpretations is correct. Moreover, this point can be generalised: since there are many possible interpretations of any linguistic expression, speaker intentions and other attitudes are needed to determine the meanings of expressions across the board.

Second, LLMs lack the sorts of attitudes that are required for their outputs to have meanings. This is sometimes called the *vector grounding problem* (cf. Mollo and Millière, 2023). Since LLMs are trained on purely textual data, they do not possess any information about the entities or sets of entities to which meaningful expressions refer. As a result, they cannot form the intention to refer to the sorts of things many of our words refer to, or interpret the communicative intentions of others (cf. Bender and Koller, 2020). Moreover, given the way that they are trained and process information, LLMs lack the sorts of attitudes needed to determine the meanings of their outputs.

In conclusion, if communicative intentions and other attitudes are needed to determine the meaning of an expression on a particular occasion of use, and LLMs do not possess these attitudes, it follows that their outputs have no meaning.

# 3   Meaning and Intention

In defence of the first premise, we begin by discussing several cases in which a single word-form may be used to refer to different entities or sets of entities, and in which a single sentence-form may be used to express different propositions. In these cases, it is intuitively obvious that speaker intentions are needed to determine what is referred to or what is being said. We then go on to generalize this point, arguing that it holds across the board and not just in isolated cases.

---

[1] For the purposes of this paper, we take propositions to be sets of possible worlds, but nothing we say hinges on this assumption.

### 3.1 Ambiguity

Let's start by considering cases involving lexical and structural ambiguity. Suppose Sita asserts the sentence in (1) or (2).

(1) A bat is stuck in the window.

(2) Frank hit the boy with the briefcase.

Starting with (1), this sentence may be used to express two distinct propositions, since it contains the lexically ambiguous word-form 'bat', which can be used to refer to an animal belonging to a species of flying mammals, call this *bat$_m$*, or to a type of equipment used in various sports such as cricket and baseball, call this *bat$_s$*. A wordform is lexically ambiguous when it can be used to refer to at least two unrelated things, and where the two meanings associated with the word-form are not even indirectly related. In other words, a lexical ambiguity is essentially a kind of linguistic coincidence, where a single word-form just happens to refer to two different kinds of things.

It is standardly assumed that in a language with ambiguous word-forms, the lexicon of the language contains two distinct words which just happen to be *homonyms*, words that are orthographically and phonologically identical. In the case of *bat$_m$* and *bat$_s$*, both words are also count nouns, which means that they have identical syntactic distributions. Since 'bat' is ambiguous, (1) could be used to literally express at least two truth-conditionally distinct propositions: that *a bat$_m$ is stuck in the window*, and that *a bat$_s$ is stuck in the window*.[2]

The sentence in (2) is also ambiguous: it can be used literally to express two truth-conditionally distinct propositions, the proposition that *Frank used the briefcase to hit the boy* or the proposition that *Frank hit the boy who was holding the briefcase*. However, in this type of case, the ambiguity is not lexical, but rather structural.[3]

Now, given that the propositions that can be expressed by (1) and (2) respectively are truth-conditionally distinct, the key question is this: When Sita asserts either (1) or (2), what has Sita literally said?

The obvious answer, we maintain, is "it depends". Typically, if Sita is sincere, then when she asserts (1), she does so with the intention of expressing some belief, which will involve either the concept BAT$_m$ or BAT$_s$, which either refers to a species of flying mammal or to a type of sporting equipment. Which proposition Sita expresses seems to clearly depend on her communicative intentions. In particular, if Sita intended to convey the proposition that *a bat$_m$ is stuck in the window*, and as a result asserts (1), then it seems quite natural to say that what Sita has literally said on that occasion is that a flying mammal is stuck in the window rather than a piece of sporting equipment.

---

[2] We adopt the convention of using italics to denote meanings or contents, and small caps to denote concepts and thoughts, which we take to be mental representational states that *have* contents.

[3] Technically speaking, at LF, the prepositional phrase 'with the briefcase' either functions as the sister node of the noun 'boy' (which would generate the second meaning) or as the sister node of the verb phrase 'hit the boy' (corresponding to the first meaning).

It is worth noting that 'bat' could refer to a variety of other things as well, such as '.bat files', the verb 'to bat', and so forth. But for the purposes of this argument, we only need to consider two distinct meanings. In other words, at least one aspect of the meaning of Sita's assertion appears to be fully determined by her referential intentions.

Now, Bender and Koller (2020) have argued that LLMs cannot have such referential or communicative intentions because they have no independent causal contact with bats or briefcases, and so cannot intend to refer to them. One prominent response to this line of argument, which we discuss below, is that LLMs do not need to have direct causal contact with objects in order to refer to them, but may refer simply by virtue of deferring to others or the linguistic community as a whole, in their use of the relevant expression (see e.g. Cappelen and Dever 2021; Mandelkern and Linzen 2024). To be sure, a speaker may be competent in the use of a name, such as 'Aristotle', without having had any causal contact with the referent of the name.

Our point here is distinct from Bender and Koller's, and holds even in cases where what Sita has literally said is merely a function of (a kind of) linguistic deference. To illustrate, consider a case where Sita overhears (1), but does not know whether the speaker intended to use *bat$_m$* or *bat$_s$*. Nevertheless, when she is later asked about what is stuck in the window, she asserts (3).

(3) Someone said that a bat is stuck in the window.

Given Sita's information state, she must be deferring to the previous speaker, so what Sita has literally reported is determined not by Sita's intentions, but by the intentions of the speaker from whom she heard that a bat is stuck in the window. Indeed, it is plausible that Sita doesn't know the literal content of her report in this case. However, it is also plausible that this kind of use of lexically ambiguous word-forms is quite rare.

By contrast, suppose Sita sees that something is stuck in the window. She believes that *it's a bat$_m$* and consequently she asserts (1). However, it turns out that it's a piece of sporting equipment. Let's assume that this is evident to everyone but Sita, so Sita's interlocutors understandably interpret her as having said that *a bat$_s$ is stuck in the window*. Now, ask yourself this: Is what Sita said true or false? We think most people are inclined to say that it's false. Assuming that this is correct, then it seems evident that what 'bat' means in this particular case is determined at least in part by Sita's intentions. Despite the fact that a much more natural interpretation of Sita's assertion is readily available, and that it is practically impossible for her interlocutors to rationally accept any other interpretation (on the assumption that Sita is cooperative and sincere), this does not suffice to sway our judgment about what she has said. Hence, Sita's intentions are clearly an essential determinant of meaning.

Lastly, suppose that Sita is not sure whether what is stuck in the window is a *bat$_m$* or a *bat$_s$*. However, she is confident it is one or the other. In this kind of case, notice that it would be very strange—indeed bordering on linguistically inappropriate—for Sita to flat out assert (1). It seems that insofar as Sita is not deferring in her use of the word-form 'bat', then in order to sincerely assert (1), she must intend to express either BAT$_m$ or BAT$_s$ in order for her assertion to be linguistically appropriate.

## 3.2 Interpretative Uncertainty

The points above generalize to various other cases of what we call 'interpretative uncertainty'—cases in which sentences can be used to literally express different contents on different occasions, where we have fairly robust judgments that which content is expressed depends on the speaker's intentions.

For example, problems analogous to those arising with ambiguity also arise in cases involving context-sensitive expressions and anaphora, where the resolution of these elements intuitively depends on the speaker's intentions. Let's consider some examples.

(4)  Sita dropped the plate. It shattered.

On the most natural interpretation, the pronoun 'it' in (4) inherits its meaning from the phrase 'the plate' in the preceding sentence. Pronouns whose meanings are determined by a linguistic antecedent are called *anaphoric* pronouns. So, what 'it' intuitively means in (4) is simply 'the plate'. However, even though many pronouns are naturally interpreted as anaphoric, pronouns generally have non-anaphoric interpretations too. For example,

(5)  Sita told David a joke. He laughed uncontrollably.

In (5), the pronoun 'he' in the second sentence is also naturally interpreted as *anaphoric*, i.e. as picking up its reference from the proper name 'David' in the preceding sentence. However, it can also be interpreted as non-anaphoric (also called *deictic* or *demonstrative*), if, for example, the speaker is demonstrating a relevant referent (by pointing, for instance). In that case, 'he' will intuitively refer to whoever the speaker is demonstrating, rather than David.

In cases such as (5), it again seems clear that what is literally said intuitively depends on the speaker's intentions. If the speaker intends an anaphoric interpretation, the meaning of 'he' is simply David and if the speaker intends a non-anaphoric interpretation, the meaning of 'he' is whoever the speaker is demonstrating. So, if an LLM were to output a token such as (5), there would arguably be no fact of the matter as to what 'he' means, if an LLM has no relevant intentions.

Of course, one could respond here that since the non-anaphoric interpretation of the pronoun in (5) requires a demonstration, and LLMs have no means of demonstrating anything, only anaphoric interpretations are available. But one immediate problem with this argument is that non-anaphoric uses of pronouns do not strictly require demonstrations. A pronoun can be used non-anaphorically without pointing at, or in other ways demonstrating, the intended referent. For example, in cases where the intended referent is sufficiently contextually salient, non-anaphoric pronouns can often be used without any demonstration being necessary.

Alternatively, perhaps one could argue that LLMs are simply linguistic agents with some inherent limitations. For example, since they cannot survey their immediate physical environment, they can only use pronouns anaphorically. Unfortunately, it is easy to demonstrate that this does not solve the problem either. Consider the sentence below.

(6) Sita saw Maya from a distance. She was wearing her red glasses.

Set aside the non-anaphoric interpretations of the pronouns in (6). The most natural anaphoric interpretation is perhaps that 'she' and 'her' (in 'her glasses') refer to Maya. However, this interpretation is by no means mandatory. There is an alternative interpretation of the sentence available where 'she' and 'her' refer to Sita. That interpretation may be slightly less natural, but with a bit of contextual priming, it is easy to get. Just imagine, as background, that Sita has very poor eyesight and can only reliably see things from a distance when she wears her red glasses. In this case, it is easy to interpret 'she' and 'her' in the second sentence as referring to Sita. Indeed, there are several other interpretations possible here that become very natural with a bit of contextual priming, such as the interpretation where Maya is wearing Sita's red glasses, or vice versa.

This now raises the question: what, if anything, determines what is literally said by an assertion of (6)? To the extent that we accept the thesis that a speaker can express a single, unique proposition when asserting a sentence with multiple possible meanings, the most natural answer—really the only plausible answer—is that it is the speaker's intentions that determine the proposition that is asserted. Given this, it seems perfectly clear that if an LLM produced a sequence of tokens corresponding to (6), without having the relevant intentions to determine a unique proposition, there simply would be no fact of the matter as to what was literally said. Hence, if an LLM were to output (6), it wouldn't have asserted anything. The token output would, strictly speaking, be meaningless.

**3.3   Context Sensitivity**

The problem described above concerning anaphora is, of course, a subset of a more general problem concerning context-sensitive expressions. When speakers use context sensitive expressions, what is literally said intuitively depends on the speaker's intentions in many cases. Consider the following example.

(7) Malik is quite tall.

Suppose that Sita asserts (7), and that she is referring to the 5-year-old boy, Malik, whom she babysits, and who is 130cm tall. Given that the average height of a 5-year-old boy is 102-109cm (according to ChatGPT), Sita's assertion is intuitively true. However, this of course requires that we interpret the predicate 'is quite tall' in a very specific way: as a claim about height relative to some specific comparison class. Obviously, Malik is not quite tall for a human being. In fact, Malik is quite short, compared to the average height of a human being. So, on the assumption that Sita said something that is literally true, what she said must be something which makes reference to a class of individuals where the average height is well below 130cm, e.g. the class of 5-year-old boys.

Of course, what exact form the semantics for gradable adjectives such as 'tall' should take is a complicated question, but that need not worry us here. All we need to observe is that 'tall' is a context-sensitive term and that it is the speaker's intention that generally resolves the interpretation of this context-sensitive element. This resolution in turn determines the truth conditions, and hence the meaning, of the assertion of (7) in context. So, insofar as LLMs have no such intentions, it remains indeterminate what is actually said when an LLM outputs a token sentence involving the word 'tall' – assuming that it is not piggy-backing on an antecedent user prompt.

It should be fairly easy to see that examples of the kind above are extremely easy to replicate for a whole range of other context-sensitive terms, such as quantificational determiners, predicates of taste, modals, counterfactual conditionals, various propositional attitude verbs, etc.

### 3.4    Interpretative Uncertainty Generalised

At this point, we suspect that some readers might be inclined to think that the above observations are insufficient to establish the very general conclusion of the argument at the beginning of this paper. After all, if intentions are needed merely to resolve ambiguities or settle cases of interpretative uncertainty, does this really warrant the sweeping conclusion that *all* the outputs of LLMs are meaningless? Perhaps there are at least some terms that are not ambiguous or context-sensitive, and thus may be meaningfully output by an LLM even without being accompanied by any referential or communicative intentions. However, the problem observed above is not limited to cases such as those we have canvassed, in which the context-invariant *standing* meaning of an expression permits it to be used to express different propositions on different *occasions of use*. We now turn to a more general argument for the claim that speaker intentions, and other attitudes, are needed to determine the correct interpretation of any use of an expression, and thus for any expression to have a standing meaning in the first place.

First, in and of themselves, expressions are just physical objects or events, such as phonetic strings, sequences of characters, or physical gestures (in the case of sign languages). They have no significance on their own. That is, the fact that an expression has a meaning in e.g. English is not grounded directly in the intrinsic phonological or orthographical properties of English expressions. Rather, the set of English expressions, individuated by their phonological or orthographic properties, could be interpreted in a multitude of different ways. Indeed, as Lewis (1956) points out, there are infinitely many possible languages that could be spoken, and thus infinitely many possible interpretations of any set of sounds or marks.

Second, the standing meanings of expressions in a public language such as English are determined by *linguistic conventions*. And just as many conventions are arbitrary––such as the convention of eating with chopsticks or one's right hand––so too is it arbitrary which conventions govern the standing meaning of an expression at a particular time and place. This should be obvious, given that the standing meanings of expressions are not fixed once and for all, but vary from one time and place to the next. There are countless examples of expressions that have changed their meanings over time, such as 'awful' and 'bully', and expressions that have different meanings in different speech communities, such as 'boot' or 'football' in British as opposed to Canadian English. Moreover, if we widen our gaze to take in all the conventions that *could* govern the use of any set of expressions, we quickly realize that the possibilities are endless. If the Angles and Saxons had died out before journeying to the British Isles, the English language might have developed differently, or might never have been spoken at all.

Third, as Lewis (1956) also notes, public languages function as a means of communication among members of a linguistic community. However, in order for languages to serve this purpose, members of the linguistic community must reliably converge on a shared interpretation of linguistic expressions on particular occasions of use. After all, communication succeeds only if the audience understands what the speaker intends to communicate.

Fourth, communicative intentions are necessary to determine *which* conventions govern the use of a linguistic expression on a particular occasion of use. For instance, suppose that Sita asserts the following sentence:

(8) There is something in the boot.

Given that the term 'boot' refers to footwear in Canadian English, but may be used to refer to the storage space at the back of a car in British English, Sita's intention is necessary to determine which variant of English she is speaking.

Of course, as a general rule, the communicative intention of the speaker does not *suffice* to determine the meaning of the expression she uses. One cannot, like Humpty Dumpty, use a word to mean just what one chooses it to mean (Carroll, 1871). Rather, speakers' intentions are constrained by the conventions of the linguistic community in which the speaker participates. In cases of interpretative uncertainty, it is plausible that a speaker cannot intend to express a proposition that requires a disambiguation or a resolution of context sensitivity that the speaker has no reasons to think that the audience will be capable of performing. For example, Sita cannot assert (8) in Toronto, among Canadians, with the intention of referring to the storage space of a car. The reason is that she has no grounds for believing that her audience is in any position to recognize this intention and, moreover, she has every reason to believe that her audience will misinterpret what she said. For this reason, Sita's assertion of (8) in that context might also intuitively come across as linguistically inappropriate, if not misleading, or even a lie.

More generally, in order to use an expression of a language with a determinate, conventional meaning, it is not enough that a speaker intends to use it to communicate something or other; rather, to attach a conventional meaning to an expression the speaker must somehow latch on to a particular set of linguistic conventions. In acts of communication, as in the cases discussed above, the speaker needs to believe that her audience is likely to understand her utterance to mean what she takes it to mean. Even when writing, with no specific audience in mind, the writer needs to believe that the words she is using have a certain conventionally determined meaning. Though the details may vary from case to case, for a speaker to use *e* to express a proposition, *p* in some language, *L*, she needs to have some beliefs about what *e* means in *L*, some desire or intention to express *p*, and some beliefs about how *e* is likely to be interpreted by others, or her future self.

At this point, you might be concerned that we are over-intellectualizing the use of language. Surely, we do not perform a kind of mental ceremony every time we use an expression, in which we solemnly vow to abide by some set of linguistic conventions. Indeed, most ordinary speakers of a language would be hard-pressed to state any of the conventions governing the use of expressions in the language, despite being perfectly competent in their use. Moreover, a child may be fully competent in the use of many expressions without mastering the concept of a convention or of meaning.

So, how could it be that for an expression to have a meaning, a speaker must have such intellectually sophisticated attitudes?

We sympathise with these concerns. However, they arise only if communicative intentions and beliefs about what expressions mean are assumed to be *explicit attitudes*: thoughts that are composed of concepts, have propositional contents, and can be expressed linguistically. If one has an explicit intention to communicate that *p*, one must have the concept of communication in addition to all of the concepts needed to think that *p*; and if one has an explicit belief that using *e* to express *p* is permitted by the conventions of a language, then one must have mastered concepts of expression, permission, and convention.

Implicit attitudes, in contrast, encode information non-conceptually, lack propositional contents, may be inexpressible in language, but are nonetheless manifest in behaviour, (cf. Nosek and Banaji).[4] Implicit attitudes are implicated in a wide array of our cognitive abilities, since there is substantial evidence that these abilities are preserved even when explicit memory and learning are impaired, see e.g. Squire, Bayley, Smith, et al. (2009).

Implicit beliefs encode information about a target system by containing internal representational units that correspond to the entities of the system, and whose properties systematically correlate with the properties of the target entities to which they correspond, cf. Burge (2018). That is, a speaker's implicit beliefs about what expressions mean are encoded in a system of concepts, which stand in a one-to-one correspondence with expressions in the language she understands, and which model the semantic properties of the expressions to which they correspond, such as their semantic types, and the contributions they make to the truth conditions of the sentences in which they occur (cf. Evans 1981, 1982; Davies 1987, 1989). Implicit intentions amount to connections from internal representations to output behaviour.[5] For instance, a speaker's communicative intention to assert the proposition that *p* is encoded in the fact that her thought that *p* explains her linguistic behaviour.

To illustrate, consider Sita once again. As an English speaker, she has acquired an array of concepts, corresponding to each of the English expressions that she understands. Given that she understands both meanings of the English term 'bat', for instance, she has two concepts, BAT$_m$ and BAT$_s$, which mean *bat$_m$* and *bat$_s$*, respectively, and which she can combine with other concepts to form explicit attitudes. For instance, when Sita forms the belief that a bat is stuck in the window, her belief will be composed of one of these two concepts. Now, suppose that this is BAT$_m$. If her having this belief leads her to assert (1), what she intends to communicate is the proposition that *there is a bat$_m$ stuck in the window*, and her use of the term 'bat' refers to a species of mammal, rather than a baseball bat.

More generally, Sita's implicit beliefs about what expressions of her language mean are encoded in the one-to-one correspondence between her concepts and the linguistic expressions to which they correspond, while her implicit communicative intentions are encoded in the role these concepts play in explaining her linguistic behaviour and dispositions.

---

[4] Implicit attitudes are often said to be inaccessible to consciousness (Nosek and Banaji; Squire and Zola). However, because implicit attitudes are typically inexpressible in language, it is difficult to distinguish between those implicit attitudes that are accessible to conscsciousness but ineffable, from those that are truly inacessible to consciousness and thus sub-personal, such as neurological processes in the brain.

[5] Strictly speaking, implicit attitudes are realized in internal representational systems which have both directive and descriptive roles, cf. Sterelny (2003). Nonetheless, it is possible to abstract from this structure to distinguish these roles.

## 4  Implicit Attitudes in LLMs

Much like expressions of a language, the outputs of LLMs are physical objects—sequences of characters or bits—which have no intrinsic meaning, and could be interpreted in a multitude of different ways. So, much like expressions, the outputs of LLMs are meaningless unless they are produced with the right kinds of implicit intentions, beliefs or desires. However, given the way they are structured, the way they process information, and the way they are trained, they cannot have implicit attitudes of a suitable kind.

The first obstacle that LLMs face derives from the fact that information about the communicative functions of text is lost in the process of encoding, so there is a sense in which LLMs cannot know, even implicitly, that their inputs and outputs encode expressions of a language. This is because LLMs do not directly process textual inputs; rather, the text that is input to the LLM must first be broken down into smaller units, or 'tokens', which are assigned a unique numerical ID stored in a token library. And since the tokenising process is optimised for efficiency in processing, not for capturing the semantic properties of expressions, information that is necessary for semantic interpretation is lost. For example, in Byte Pair Encoding, which is used by many LLMs, the tokeniser begins with a base library consisting of a set of characters (such as ASCII or Unicode characters), and performs a sequence of merge operations which result in the addition of short sequences of characters that occur frequently in the input text to the token library. So, if the tokenizer is trained on English text, the token library is likely to include frequently occurring sequences such as 'ee', and 'ea', but not 'aa', since 'aa' occurs infrequently in English. Since the most frequently occurring strings of characters are not typically the smallest units of significance, such as words or morphemes, the tokens do not encode information about which expressions are basic units of meaning (Millière and Buckner 2024). Thus, even the minimal information about which units are meaningful is lost in the tokenising process.

Furthermore, since each token receives a unique numerical ID, the vector encoding of a sequence of tokens does not vary depending on what it means, and thus cannot resolve ambiguity or interpretative uncertainty. The sequence of characters 'b', 'a', 't', for instance, will be encoded in the same vector, $v_{bat}$, regardless of whether or not it is used to refer to a mammal or to an item of sporting equipment (Lake and Murphy 2023). So, even if an LLM were to output (1) and was caused to do so by some internal state involving $v_{bat}$, this would not suffice to disambiguate the LLM's output of 'bat'. Indeed, *each* of the sentences in (1) to (7) will be encoded in a unique vector, irrespective of the context of use, and thus irrespective of meaning. So, even if we suppose that the vectors that cause the LLM's outputs figure in the contents of its implicit intentions, they do not suffice to disambiguate expressions or resolve interpretative uncertainty. More generally, since a set of vectors could be interpreted in a multitude of different ways, even if an LLM's token library were to include all and only the lexical atoms of a language, that would still leave infinitely many interpretations of those vectors open.

Indeed, there is reason to doubt that an LLM has any way of encoding the information that its vectors represent linguistic expressions, let alone what those expressions mean. After all, an LLM is just a foundation model, often with a transformer architecture (Vaswani et al. 2017), which happens to have been trained on linguistic data. But foundation models can be trained on a wide variety of different types of data: images (Zamir et al. 2022), brain scans (Tang et al. 2023), protein folding (Jumper et al. 2021), and more (Islam et al. 2024). In each application, the data is tokenised before it is encoded in vectors, and it is really only the vectors which are operated on by the transformer. These vectors encode information about sequences of numerical token IDs, without encoding any information about what data the tokens encode. A transformer has no way of knowing whether its vectors encode linguistic data, sequences of amino acids, a user's browser history, or the brain activity of a fish.

Another reason to think LLMs lack implicit beliefs about what expressions mean has to do with how they are trained. During pre-training, the LLM receives inputs in the form of vector encodings of incomplete sequences of text drawn from its training data, performs a series of mathematical operations on those vectors, and outputs a probability distribution over the tokens in its library, from which it samples an output. These mathematical functions calculate the probabilistic correlation between any two pairs of vectors in the input stream, allowing the system to model long-range statistical dependencies in the data more accurately than previous models. The LLM is then shown the token that in fact completes the sequence in the dataset, calculates a loss function to minimize its prediction error, and then adjusts the weights by resetting parameters in the functions it computes. Fine-tuning proceeds in much the same way, only on more specialized datasets, and may involve human feedback to determine the 'correct' completion of a sequence. So, even when an LLM is trained to answer questions, it is trained to *predict* the sequence of tokens (the 'answer') that is most likely to follow the input sequence (the 'question') in its dataset, without knowing that what the input vectors encode are linguistic expressions that may be used to ask questions, or that its outputs are answers to those questions. That is, from start to finish, LLMs are trained to *predict tokens*, so if they have any implicit beliefs at all, they are beliefs about the statistical correlations between tokens in the datasets on which they are trained, and if they have implicit intentions, they intend to predict the masked token in a sequence of tokens.

Finally, the way LLMs process information contrasts starkly with the way in which information is processed in human language users, suggesting that they lack implicit semantic beliefs or communicative intentions. For instance, suppose we ask Sita:

(9) How can I get from Jackson Heights to NYU?

And she answers:

(10) You can take the F from Roosevelt Avenue and get off at West 4$^{th}$ St.

What goes through Sita's mind? Presumably, when she hears (9), she interprets the question in light of her implicit semantic beliefs, such as that 'Jackson Heights' refers to a neighbourhood in Queens, and that 'NYU' refers to New York University (more specifically, in the context, the Greenwich Village Campus). According to the model of implicit belief sketched above, this amounts to Sita entertaining a thought containing concepts JACKSON HEIGHTS and NYU, which correspond to the English names, 'Jackson Heights' and 'NYU', respectively. Having interpreted (9), she has an internal representation of the possible answers to the question, the set of possible ways of getting from Jackson Heights to NYU. She then consults her memory and recovers the explicit beliefs, such as THE F TRAIN FROM ROOSEVELT AVENUE TAKES YOU TO WEST $4^{th}$ STREET STATION, and WEST $4^{th}$ STREET STATION IS NEAR NYU. Then, these beliefs, along with her desire to answer the question in (9) sincerely, lead her to assert (10), thereby constituting her implicit intention to communicate the contents of her beliefs.

This process is nothing like what goes on in the internal workings of an LLM. Suppose we input (9) into an LLM. It will first break (9) down into tokens, 'H', 'o', 'w', and so forth, and encode the sequence of tokens in a vector. It then calculates the 'relatedness' of each of the tokens encoded by the input vector with each of the tokens stored in its vector space, and outputs a probability distribution over the tokens in its library, from which it selects the next token, which may be no more than a character, such as 'Y'. Since many LLMs are auto-regressive models, they take the output token and append it to the input vector, and go through the process again (Wolfram 2023). When the LLM consults its memory, it remembers stored information about sequences of vector encodings of tokens, such as,

$$v_{Ja}, v_{ck}, v_{son},$$

and the likelihood that this sequence of tokens will co-occur with each of the other tokens in its vector library. Crucially, at no point in this process does the LLM rely on implicit beliefs about what expressions mean to interpret the question, nor implicitly intend to express the contents of its beliefs about Jackson Heights or NYU in outputting (10). Given the way in which it has been trained, the LLM simply does not have any memories about Jackson Heights, NYU, or the way to get from one place to the other to consult.

Indeed, there is some degree of randomness in the LLM's selection of an output token, which is controlled by a 'temperature' parameter that is set by an engineer to optimise the LLM's performance in connection with a particular type of application. When the temperature is set too low, the outputs of the LLM tend to be sensible but stilted and unnatural, and when it is set too high, its outputs make little sense. (Interestingly, the sweet spot for LLMs is around 0.7 or 0.8 (Wolfram 2023).) The fact that an LLM's choice of output token is random in this way represents a stark contrast with the process taking Sita from hearing (9) to asserting (10). Though Sita may have many options regarding which words to use to express her beliefs, and her choice is free, the expressions she samples from have one thing in common: they can be used to express the same propositions. In contrast, the set of tokens from which the LLM samples an output do not have this in common. First, given that tokens do not correspond to the smallest units of significance in a language, some of the tokens from which the LLM samples an output have no meaning at all. Second, the auto-regressive processing employed by many LLMs makes their outputs path-dependent, in the sense that the LLM's selection of the $n^{th}$ output token is constrained by its previous selections. As a result, as the temperature increases, so too does the variability of the outputs an LLM is disposed to give to one and the same input. And any time the temperature is non-zero, the path-dependence of the LLM's outputs makes it possible for it to output semantically heterogeneous, even inconsistent responses to the same question on different occasions, in a way that cannot be accounted for by any kind of sensitivity to context. This heterogeneity is evident in the plethora of cases in which LLMs vary between outputting sentences that are true on their most natural interpretation, and outputting 'hallucinations'—sentences that are false, on their most natural interpretation (Reddy, Pavan Kumar, and Prakash 2024).

## 5 Resisting the argument

There are several ways in which one might resist the foregoing argument, either by challenging the premises or by challenging one of the background assumptions that we make while defending the premises. We consider some of these potential responses to our argument below.

### 5.1 Semantic Externalism

One interesting challenge to the first premise of our argument, and thus to its conclusion, derives from the tradition commonly known as *semantic externalism*, according to which the reference of an expression is fully determined by facts that are external to the goings-on in the mind of a language user (cf. Cappelen and Dever 2021) and Mandelkern and Linzen, 2024). According to externalists, it is possible for the use of an expression to have a meaning even if its producer is ignorant of all of the external facts that determine what it means.

To begin with, consider Mandelkern and Linzen's (2024) argument that speaker intentions are not required for an expression to have a meaning, only that the speaker interacts with a linguistic community, and that LLMs plausibly satisfy this requirement. To illustrate, consider the following example, adapted from Mandelkern and Linzen (2024). Suppose Lucy stumbles across a (rather unreliable) mathematics webpage which states that Peano proved the incompleteness of arithmetic. Suppose further that, prior to encountering this page, she had never heard of Peano, had absolutely no discriminatory capacities with regard to identifying Peano, and no beliefs about Peano. After reading the webpage, Lucy comes to believe that Peano proved the incompleteness of arithmetic. Later, she encounters her math teacher and says,

(11) I learned something interesting yesterday: Peano proved the incompleteness of arithmetic.

In this case, it is natural to interpret Lucy as saying something false, since it was Gödel who proved the incompleteness of arithmetic, but nonetheless saying something about Peano, and thus successfully referring to the Italian mathematician, Giuseppe Peano. Mandelkern and Linzen maintain that Lucy can use 'Peano' to refer to Peano in virtue of the fact that she acquired the name by interacting with a speech community in which 'Peano' was used to refer to Peano, even though she has never had any direct causal contact with Peano, and though her only belief about Peano, the referent of 'Peano', is false (Mandelkern and Linzen, 2024, 23).

Similarly, assuming externalism, Mandelkern and Linzen argue that an LLM would not need to know anything about the natures of the entities in the external world to which a linguistic expression refers in order to use that expression to refer to them. Nor would the successful use of an expression to refer to some set of entities require that the LLM has any particular beliefs or presuppositions about those entities, or dispositions to discriminate those entities from other, similar ones. Since meaningful use is compatible with ignorance about the determinants of meaning, they argue, the outputs of an LLM could be meaningful even if it did not encode any of this meta-semantic information. Rather, they continue, it suffices for a use of an expression to have a meaning that the user of the expression defers to a linguistic community in which it is used with that meaning, by virtue of having acquired the expression through interaction with that community. We call this weak deference:

> *Weak Deference.* A use of an expression *e* has its conventionally determined meaning if the user of *e* acquires it by interacting with a linguistic community in which *e* has a history of being used with that extension.

Mandelkern and Linzen go on to argue that there are no convincing reasons to think that an LLM could not interact with a linguistic community and thereby come to use expressions with the meanings they have acquired through their history of use in that community. Since the data on which an LLM is trained encodes expressions that have a history of meaningful use in a linguistic community, an LLM can subsequently use those expressions with their conventionally determined meanings. They say (Mandelkern and Linzen, 2024, 4):

> LMs obviously have the right kind of grounding to refer: if they are part of a linguistic community which uses 'Peano' to refer to Peano, then their use of 'Peano' refers to Peano. [...] One way to formulate the worry about grounding, again, is in terms of form grounding meaning: LMs have access only to form, and form underdetermines reference. Now we can see what is wrong with this argument. The inputs to LMs are not just forms, but forms with particular histories of meaningful use. And those histories suffice to ground the referents of those forms.

Though we agree with the externalist premise underlying Mandelkern and Linzen's argument, we remain sceptical about their claim that speakers' intentions play no role in determining reference. First, it is worth noting that Mandelkern and Linzen's view is not a direct implication of externalism, since one can use an expression with the intention to refer to *x* or express the proposition that *p*, without knowing anything about the nature of *x*, or how its reference is determined. Indeed, according to Kripke, one of the founding fathers of externalism, a *deferential intention* is needed for a language user to 'borrow' the referent of an expression from another, or the linguistic community in which it has a history of meaningful use. For instance, Kripke wrote that when a name is acquired by deference (Kripke, 96),

> [...] the receiver of the name must...intend when he learns it to use it with the same reference as the man from whom he heard it. If I hear the name

'Napoleon' and decide it would be a nice name for my pet aardvark, I do not satisfy that condition.

Kripke's point is that one defers in one's use of an expression only if one intends to defer, because otherwise, one's expression does not acquire a derived reference, but either a new one, or none at all. Mandelkern and Linzen anticipate this point, arguing that for the view that communicative intentions are needed to determine reference to be remotely plausible (Mandelkern and Linzen, 6):

> [...] it must be that this kind of intention is pretty lightweight: it must be sub-personal (many language users would not articulate this kind of intention if you asked them) and it must be insubstantial in the sense that it must take the form of an intention to generally be part of a linguistic community, rather than to refer to whoever satisfies such-and-such properties.

Thus, according to Mandelkern and Linzen, the question is really whether LLMs can have lightweight intentions of this kind, and though they acknowledge having no clear argument that LLMs can have such intentions, they don't see a clear argument that they can't.

Our response to the foregoing is twofold. First, we dispute Mandelkern and Linzen's claim that any referential or communicative intention would have to amount to a general intention to be part of a linguistic community. Though such general intentions may be necessary to determine which language is being spoken, we have argued that there are also many cases in which specific referential or communicative intentions are needed to resolve ambiguities or settle interpretative uncertainties.

Second, though it is not entirely clear whether Mandelkern and Linzen's use of 'lightweight' aligns perfectly with ours, the implicit attitudes that we claim are necessary to determine meaning are lightweight in the sense that a language user may well have an implicit referential or communicative intention without being able to articulate it or report it in a language.[6] For instance, the fact that Lucy implicitly defers to the author of the webpage in her use of 'Peano' does not require that she have any explicit beliefs about that person. Rather, her deference is grounded in the fact that she acquired the concept Peano as a result of reading that page, and that Peano may be combined with other concepts of hers to formulate explicit attitudes, whose contents she is disposed to express by using the name 'Peano'. If Lucy were to later decide that 'Peano' would be a nice name for her pet aardvark, she would have to acquire a distinct concept, Peano*, which would explain her uses of the name of her pet aardvark. As we have argued above, LLMs cannot have implicit intentions such as these, and so they cannot produce outputs with a meaning that derives from their history of meaningful use.

Cappelen and Dever's (2021) line of objection to our first premise similarly rests in part on the assumption of semantic externalism. They begin with the methodological argument that when interpreting AI systems, we cannot simply assume that theories devised for humans may be appropriately applied to artificial systems. They argue that when semantic externalism is appropriately 'de-anthropocentrised', meaning and reference do not require any attitudes.

---

[6] We do not take implicit intentions to be *sub-personal*, as the notion was defined by Dennett (1969), to refer to processes that are wholly inaccessible to consciousness, like digestion. It is not clear, however, whether Mandelkern and Linzen intend to use 'sub-personal' with this meaning.

The view they propose is a form of *pure* externalism, according to which meaning and reference are determined exclusively by causal relations to the external environment (see also Cappelen and Deutsch, 2024). For instance, the meaning of 'Peano', when Lucy uses it, is determined by a causal chain linking Lucy's use of the name, via the author of the webpage, back to Peano, at the moment when he was baptized. They combine this account with the suggestion that names are associated with mental files (Recanati, 2012), containing the information that one associates with the name. If Lucy goes on to acquire further beliefs about Peano, such as that he was Italian, was a mathematician, etc., these get added to her mental file. However, the contents of the file do not determine the meaning of the name 'Peano'. Rather, Cappelen and Dever argue, the referent of her uses of 'Peano' is whatever it is that is the dominant causal source of the information contained in her mental file. Assuming that this is Peano, the information contained in her mental file may well be false. In place of the claim that reference requires attitudes, they propose the following principle (Cappelen and Dever, 2021, 111):

> *Denotation through dominant causal source:* A system can denote an object that is the dominant causal source of a set of information given as input in the training stage.

This implies that an LLM's output *x* can relate to an extension purely in virtue of the fact that the members of the extension constitute the dominant causal source of the set of information that is associated with *x* in the LLM's mental file. Since there is no reason to think that an LLM cannot stand in causal relations of this kind, their view implies that the outputs of LLMs are, at least in some cases, meaningful.

Though Cappelen and Dever's argument may seem superficially promising, it ultimately does not support the conclusion that the outputs of LLMs are meaningful. Here's why. On the face of it, for a system to acquire information, it must encode that information in some way. In Lucy's case, it is natural to say that the information is encoded in her explicit beliefs involving the concept PEANO. However, as all of the foregoing arguments show, LLMs cannot store semantic information either implicitly or explicitly. They cannot store information in explicit beliefs because they cannot acquire concepts on the basis of the data on which they are trained. At any rate, in the absence of an explanation of how LLMs achieve this cognitive feat, we have no reason to think that they do.

Cappelen and Dever might argue that the information that is associated with *x* in an LLM is implicitly encoded in the set of vectors that contain a vector representation of *x* as a proper part. However, for reasons that we have canvassed above, LLMs cannot have even implicit beliefs about the individuals denoted by names. This is because the vectors they process do not encode the information about the semantic functions of tokens, let alone that some sequences of tokens refer to objects in the world. But if LLMs cannot store information about external objects explicitly or implicitly, they cannot have implicit beliefs about them, and thus have no information to assemble in even a metaphorical mental file.

Perhaps Cappelen and Dever will be inclined to argue that the meanings of an LLM's outputs are determined by the dominant causal source of the expressions in the data on which the LLM is trained, without the LLM needing to encode any information about the dominant causal source itself. However, given that the tokens LLMs process do not correspond to expressions, it is not clear that all tokens could inherit the dominant causal source of an expression, in the absence of information with the dominant causal source. Consider, for instance, the case of homonymous names, such as 'Muhammad'. Though there are millions of people called Muhammad, an LLM will have just one vector encoding of the name. If the data on which the LLM is trained contains many occurrences of a name, with different causal sources, there may be no dominant causal source of the vector representation of it. Or if it just so happens that a famous boxer is the dominant causal source of the vector representation, then any time it outputs the name—perhaps when telling some other Muhammad his credit score—it will refer to the famous boxer.

A deeper worry about this proposal is that if the vectors processed by the LLM are not assumed to carry information, the mere fact that some vector has a dominant causal source does not suffice for it to be a representation. After all, there are causal relations all over the place, but not every effect of a cause is a representation of its cause.

So, semantic externalism does not seem to provide grounds to block the first premise of our argument.

### 5.2 Semantic Internalism

One way to respond to our argument would be to reject two of our background assumptions: that the meanings of expressions in a public language such as English fundamentally involve relations to extensions—the entity, or the set of entities that the expression represents; and that the literal standing meaning of an expression is determined by the conventions of a linguistic community. Though these assumptions are commonly made among semanticists and philosophers of language, there is a family of approaches to linguistic meaning which reject them. According to *semantic internalists*, an expression derives its meaning from features of the concept that an individual associates with it, which are internal to her mind. For instance, according to conceptual role semanticists, the meaning of a word is determined by the role played by the associated concept in an individual's reasoning, or in guiding her actions, cf. Block (1986), Chalmers (2021), Loar (1981, 1988), Harman (1999), Greenberg and Harman (2005).[7]

One influential response to Bender and Koller's (2020) objections appeals to conceptual role semantics to bolster the claim that at least some of the outputs of LLMs are meaningful (Piantadosi and Hill, 2022).[8] The thought is that LLMs encode

---

[7] Perhaps the best-known internalist is Chomsky (2014), who maintains that the meanings of public language expressions are epiphenomena, projections of the meanings of expressions in a speaker's private language, which are heavily constrained by linguistic rules encoded in the speaker's innate language faculty. We set aside Chomsky's view here, since it seems to block any route to resisting our argument, as Chomksy 2023 himself argues.

[8] Søgaard (2022), defends a stronger form of internalism, that applies to all expressions, and argues that LLMs represent meanings on the bases of the observation that the distribution of vectors in an LLM's latent

structural information about the relations between tokens, from which they are able to recover semantic information about them, without requiring any independent access to any of the entities their expressions are about, cf. Piantadosi (2023). Conceptual role semantics is most naturally applied to logical terms, such as 'and', 'not', 'or', 'all', and 'some', since these do not refer to entities in the world Piantadosi and Hill (2022). For instance, according to many proponents of conceptual role semantics, the meaning of the word 'and' is determined by the introduction and elimination rules one might learn in a first year introductory course in logic:

&-Introduction (&I):
$$\frac{A \quad B}{A\&B}$$

&-Elimination (&E):
$$\frac{A\&B}{A} \quad \text{and} \quad \frac{A\&B}{B}$$

The thought is that what it is to use 'and' to mean the truth-functional connective of conjunction is to be disposed to conform to &I and &E in one's use of 'and'. So, if an LLM is disposed to conform to these rules, then when it outputs 'and', it too means conjunction.

    Let us grant that conceptual role semantics provides the correct account of the meanings of logical terms, and that the inputs and outputs of LLM's conform to &I and &E. Nonetheless, we maintain, when an LLM outputs 'and', it is devoid of meaning. The reason is that &I and &E are *schematic templates*, syntactic objects that are meaningful only in conjunction with a specification of what the dummy letters *A* and *B* are placeholders *for*. If *A* and *B* are assumed to be placeholders for propositions which may be true or false, &I and &E can be interpreted as stating that a conjunction is true if and only if its conjuncts are, and thus specifying a *conceptual role*.[9] However, Boolean operations that are structurally equivalent to &I and &E have widespread applicability outside of logic. An AND-gate, for instance, may be represented schematically as in Table 1, where the values 1 and 0 may be interpreted as representing any binary states of a system, such as on/off, bright/dim, activated/not activated, or high voltage/low voltage, depending on the application:

---

space are "near isomorphic" across different languages, perceptual spaces, and physical spaces. This view is susceptible to the same sorts of objections that we raise against Piantadosi and Hill.

[9] Some inferentialists or conceptual role semanticists might wish to formulate conceptual roles syntactically, in which case, they would take *A* and *B* to be placeholders for syntactically well-formed sentences of a language. Others might be inclined to adopt a 'deflationary' understanding of 'proposition', 'true', and 'false' (cf. Horwich, 1998). Our argument does not turn on these details.

| A | B | A&B |
|---|---|-----|
| 0 | 0 | 0 |
| 0 | 1 | 0 |
| 1 | 0 | 0 |
| 1 | 1 | 1 |

Table 1: AND-gate

Since many different systems can satisfy a Boolean structure, the fact that a system satisfies one underdetermines the correct interpretation of that structure, which may be a representational system, or an electrical panel. So, even if the inputs and outputs of an LLM conform to &I and &E, when it outputs '... and ...' it expresses the concept of conjunction *only if the blanks are filled with vectors or strings that express a propositional content*.[10] However, as we have argued, the outputs of LLMs do not express propositions. It follows that when an LLM outputs 'and', it does not express the concept of conjunction.

## 6  Attributing Meaning to LLM outputs

At this point, you might ask: if the conclusion of our argument is correct, and the outputs of LLMs are indeed meaningless, how are we to explain the fact that the outputs of LLMs *seem* to be meaningful, or the fact that they are widely used to acquire new knowledge?

The answer is, we think, quite simple. Although the outputs of LLMs are indeed meaningless, because they lack the intentions required for their outputs to be meaningful, the strings that they output are nonetheless *attributed* a meaning by us, making them *seem* to be meaningful. To illustrate with an example adapted from Putnam (1975), if an ant crawling in the sand accidentally leaves a trail resembling (1), there is no sense in which the ant has asserted a proposition in English. So, the trail has no meaning in anything but the trivial sense that any expression can be interpreted as having a meaning in some possible language or other. However, if Sita stumbles across the trail, even if she knows how it was produced, it would be quite natural for her to automatically attribute a meaning to it. If she does, she will either interpret the part of the trail that resembles the word 'bat' as referring to a type of mammal or a bit of sporting equipment, depending on whether $BAT_m$ or $BAT_s$ pops up in her mind. To attribute a meaning, she effectively *pretends* that the output was produced by a competent user of English, with the relevant explicit and implicit attitudes. But when pressed, she will readily agree that, given how it was produced, the trail *itself* has no meaning. Call meaning that is attributed on the basis of pretense *ersatz* meaning.

We contend that it is because we attribute ersatz meaning to the outputs of LLMs that they *seem* meaningful to us, despite in fact being meaningless. Moreover, in many cases, attributing ersatz meanings to the outputs of LLMs is epistemically beneficial, leading us to acquire new knowledge.[11] This is because, when we interpret any expression, we assume that the producer of the expression is rational and cooperative, and we typically rely on a good deal of background knowledge about such things as the prior linguistic context, the topic or question under discussion, and general knowledge about the world. These clues typically suffice for convergence between what a speaker intends to communicate and what the hearer interprets her as having communicated. In the case of LLMs, though there is no hope of convergence, attributing an ersatz meaning to their outputs will often maximize expected epistemic benefits, such as true beliefs or new knowledge.

---

[10] This point holds even if 'and' is defined syntactically. In that case, when 'and' figures in the LLM's outputs, it is a sentential operator only if the vectors it processes determinately represent well-formed sentences.

For example, suppose you ask an LLM chatbot the question in (11) and it outputs (12) in response.

(11) Are there any mammals that can fly?

(12) Yes, a bat is a mammal and it has the ability to fly.

As we have argued, there is simply no fact of the matter as regards what 'bat' means in (12). Nonetheless, since (12) could only be considered relevant to the question under discussion on the assumption that 'bat' denotes an animal, it is natural for us to simply assume that this is what 'bat' denotes. Another reason to select this interpretation is that it is more likely to expand our knowledge than the alternative interpretation, that 'bat' denotes a piece of sporting equipment. Though this interpretation would not, strictly speaking, be a *misunderstanding* of (12), because that would require the LLM to have communicative intentions that it does not have, it would render (12) obviously false, and thus not a good bet if one aims to acquire knowledge.[12]

Finally, some might argue that though LLMs lack communicative intentions, and thus do not express propositions, their outputs nonetheless have some form of *derived* meaning analogous to the outputs of some kind of copying device (Lederman and Mahowald 2024, Pepp, 2025). For example, if you make a photocopy of *Silent Spring*, the copy doesn't cease to have a meaning simply because the photocopier lacks communicative intentions. Similarly, Lederman and Mahowald (2024) argue, though LLMs lack communicative intentions, their outputs may have a derived meaning that results from an analogous process to that of the photocopier.

We fully agree that the outputs of photocopies are meaningful, at least typically, but only because they *preserve the meanings of their inputs*. This requires both that the inputs to the copier are meaningful, and that the process by which it produces outputs preserves input meanings. That is, the fact that your copy of *Silent Spring* is meaningful is partly grounded in the fact that it is a copy of a text that was produced by Rachel Carson, whose communicative intentions determine the meanings of the expressions in it, and partly grounded in the fact that your copier did not significantly alter the text. In contrast, if you were to take a photograph of the ant's trail, your photograph would have no more meaning than the ant's trail. Or if you were to copy out *Silent Spring* carelessly by hand, you might make such egregious errors that your copy would not preserve Carson's communicative intentions, and thus would not preserve the text's original meanings.

---

[11] This raises a number of interesting epistemological issues that we lack the space to address here. If the outputs of LLMs are indeed meaningless, and if they do not assert propositions, their outputs do not qualify as *testimony*, and therefore, not a source of testimonial knowledge. How do we acquire knowledge on the basis of reliance on LLMs, if not on the basis of their testimony? If we are mistaken in thinking that the outputs of LLMs are assertions, does this defeat any justification we may have for relying on them, even if they may provide non-testimonial knowledge?

[12] We assuming the principle that, all else being equal, we should choose the interpretation of an LLMs that has the greatest expected epistemic utility *for us*, because the aim of interpretation is not to determine what the outputs of LLMs in fact mean—since that is indeterminate—but to maximize the benefit of the pretence that they do. In contrast, Cappelen and Dever(2021) appeal to this principle of interpretation to argue that the outputs of AI systems are in fact meaningful. However, we are not entirely clear on what motivates this stronger claim.

Of course, LLMs are not photocopiers, as Lederman and Mahowald are well aware. Rather, they merely claim that there is an analogy between photocopiers and LLMs, in the sense that the outputs of LLMs have a derived meaning, which is partly grounded in the fact that the data on which they are trained is meaningful, and partly grounded in the fact that they are set up to be causally sensitive to, and therefore preserve, *intelligibility* (Lederman and Mahowald, 2024) which they define as follows (with some minor alterations for generality):

> *Intelligibility.* An expression, *e*, is intelligible in a language, *L*, if and only if it is possible for someone to understand *e* in line with the conventions of *L*.

Given this definition of intelligibility, Lederman and Mahowald may be interpreted as claiming no more than that the outputs of LLMs have an *ersatz* meaning, in which case, we agree. However, the copier analogy suggests a stronger view, according to which the outputs of LLMs have more than merely *ersatz* meanings, albeit meanings that are derived from the meanings of the expressions in the data on which they have been trained. If this is their view, we contend that it faces significant difficulties in the face of many of the points that we have raised in this paper. These can be stated in the form of a dilemma, depending on whether or not understanding is assumed to imply convergence of interpretations between the LLM and its audience.

First, suppose that understanding does not imply convergence. Then, *Intelligibility* is true, but implies no more than that the outputs of LLMs have ersatz meanings. This is because possible languages come cheap: any set of linguistic expressions can be understood in line with the conventions of infinitely many possible languages. Yet, the fact that an expression, *e*, can be understood in line with the conventions of a possible language, *L*, does not imply that the producer of *e* understands it in that way. So, if understanding does not imply convergence, then the fact that an LLM's output can be understood in line with the conventions of *L* implies no more than that the interpreter can attribute an ersatz meaning to it.

Second, suppose that understanding does imply convergence. Then, in light of the preceding arguments, *Intelligibility* is false. The reason is that meanings are dear: if linguistic expressions are not produced with the requisite communicative intentions, they are meaningless; since LLMs lack communicative intentions, there is no interpretation of their outputs with which anyone's interpretation could possibly converge. So, if understanding implies convergence, the outputs of LLMs cannot be understood by anyone and are therefore unintelligible.

Furthermore, the outputs of LLMs are not causally sensitive to intelligibility, as Lederman and Mahowald suggest, but to some non-semantic proxies, which does not come to the same thing. Indeed, this seems to be the result of applying the very counterfactual test for causal sensitivity that Lederman and Mahowald propose, provided that we control for independent variables. That is, to test whether the outputs of LLMs are causally sensitive to intelligibility, we need to consider counterfactual circumstances in which intelligibility is altered, while controlling for independent variables, such as the statistical properties of the vector representations of the data on which the LLM is trained. One such counterfactual circumstance is one in which the language of the community that produces the primary data on which the LLM is trained is not English, but English*, which has the same syntax as English, but a wildly different semantic interpretation. If intelligibility is varied in this way, while all other variables are held fixed, the LLM would plausibly learn exactly the same weights and would produce exactly the same outputs in response to prompts after it has been trained. So, it seems that LLMs are not causally sensitive to intelligibility after all.